# R-TOSS: A Framework for Real-Time Object Detection using Semi-Structured Pruning

Abhishek Balasubramaniam, Febin P Sunny and Sudeep Pasricha
Department of Electrical and Computer Engineering
Colorado State University, Fort Collins, CO, USA
{abhishek.balasubramaniam, febin.sunny, sudeep}@colostate.edu

*Abstract*—Object detectors used in autonomous vehicles can have high memory and computational overheads. In this paper, we introduce a novel semi-structured pruning framework called *R-TOSS* that overcomes the shortcomings of state-of-the-art model pruning techniques. Experimental results on the JetsonTX2 show that *R-TOSS* has a compression rate of 4.4× on the YOLOv5 object detector with a 2.15× speedup in inference time and 57.01% decrease in energy usage. *R-TOSS* also enables 2.89× compression on RetinaNet with a 1.86× speedup in inference time and 56.31% decrease in energy usage. We also demonstrate significant improvements compared to various state-of-the-art pruning techniques.

## I. INTRODUCTION

In recent years, autonomous vehicles (AVs) have received immense attention due to their potential to improve driving comfort and reduce injuries from vehicle crashes. A report from the National Highway Traffic Safety Administration (NHTSA) indicated that in 2021, more than 31,720 people were involved in fatal accidents on U.S. roadways [1]. These accidents were found to predominantly be caused by distracted drivers who contributed to ~94% of them. AVs can help mitigate human errors and avoid such accidents with the help of their superior perception systems. A perception system helps AVs understand their surroundings with the help of an array of sensors that can include Lidars, Radars, and Cameras. Object detection is an important component of such perception systems [2].

AVs must process a huge amount of data in real-time to provide precise corrections to the vehicle controller to maintain its course, speed, and direction. To assist with vehicle path planning and control, AVs rely on object detectors to provide information about the obstacles in their surroundings. These object detectors must satisfy two important conditions: 1) maintaining high accuracy, and 2) providing inference in real-time (~tens of milliseconds). In recent years researchers have been able to design machine learning models for object detection with high accuracy, but these models are generally very compute-intensive and often combined with sensor fusion task which helps in providing the input to these models by combining data from various sensors [3][35]. Apart from these object detectors Avs also must process immense data as a part of the Advance Driver Assistance System (ADAS) for operation safety and security such as in-vehicle communication and vehicle-to-x (V2X) protocols which can increase the computational cost and power usage [4][25][33]. This is a challenge because onboard computers in AVs are resource-constrained, with strict limits on power dissipation and computational capabilities.

Object detection is a compute and memory-intensive task involving both classification and regression. Typically, all machine learning based object detectors can be classified into two types: 1) Two-stage detectors and 2) Single-stage detectors. Two-stage detectors consist of a two-stage detection process that involves a region proposal stage and subsequent object classification stage. The regional proposal stage often consists of a Regional Proposal Network (RPN) which proposes several Region of Interests (ROIs) in an input image (e.g., from a camera sensor in an AV). These ROIs are used to classify objects in them. The objects are then surrounded by bounding boxes to localize them. Examples of two-stage detectors include R-CNN [5], Fast R-CNN [6], and Faster R-CNN [7]. In contrast to two-stage detectors, single-stage detectors use a single feed-forward network which involves both classification and regression to create the bounding boxes to localize objects. Single-stage detectors are lightweight and faster than two-stage detectors. Some examples of single-stage detectors are YOLOv5 [8] (You Only Look Once), RetinaNet [9], YOLOR [10], and YOLOX [11].

Unfortunately, even single-stage detectors are compute and memory intensive, so deploying and executing them on embedded and IoT boards in AVs remains a bottleneck [12]. To address this bottleneck, many techniques have been proposed in recent years, such as pruning, quantization, and knowledge distillation, to compress and optimize object detector execution, with an emphasis on improving inference time while preserving model accuracy. Pruning techniques in particular have been shown to be very effective in increasing the sparsity of object detector models, by carefully removing redundant weights that do not impact overall accuracy. Such sparse models require fewer computations, and can be compressed to reduce latency, memory, and energy costs.

In this paper, we introduce the *R-TOSS* object detector pruning framework to achieve efficient pruning of object detectors used in AVs. Unlike traditional pruning algorithms that can generally be classified as structured pruning [19]-[23] or unstructured pruning [13]-[18], we utilize a more niche approach that involves semi-structured pruning. Our approach involves applying specific kernel patterns to prune convolutional kernels and associated connectivity. The novel contributions of our proposed pruning framework for object detectors are as follows:

- An approach for reducing computational cost of iterative pruning by using depth first search to generate parent-child kernel computational graphs, to be pruned together;
- A pruning technique to prune 1×1 kernel weights to increase achievable model sparsity;
- An implementation of kernel pruning without connectivity pruning, to preserve kernel information for inference, that can help retain model accuracy;
- A detailed comparison against multiple state-of-the-art pruning approaches to showcase the effectiveness of our novel framework, in terms of mAP, latency, energy usage, and achieved sparsity.

The rest of the paper is organized as follows: Section II provides a detailed background in terms of state-of-the-art object detector models and pruning techniques; Section III describes the motivation for this work; Section IV introduces our framework and provides a deep dive into the algorithms proposed; Section V showcases our experimental results, and Section VI presents our conclusions.

## II. BACKGROUND AND RELATED WORK

### A. Object Detectors

Object detectors are used in AVs for various tasks such as traffic sign and traffic light recognition, lane recognition, vehicle tracking, etc. These object detectors can be classified into two categories, two-stage and single-stage detectors. To evaluate an object detector, irrespective of the category, mean average precision (mAP) and intersection over union (IoU) metrics are used. mAP is the mean of the ratio of precision to recall for individual object classes, with a higher value indicating a more accurate object detector. IoU measures the overlap between the predicted bounding box and the ground truth bounding box.

*Two-stage object detectors:* Two-stage detectors use a two-stage process consisting of a region proposal and object classification. R-CNN [4] was one of the first deep learning-based object detectors to be proposed. The algorithm's novelty came with an efficient selective search algorithm for ROI proposals, which dramatically decreased the overall number of regions needed to be processed. The regions were fed into a convolutional neural network (CNN) for feature extraction. The CNN output was sent to a support vector machine (SVM) for classifying the object in the region. Even though the reduction in ROI proposals was revolutionary in terms of minimizing inference, the R-CNN algorithm cannot infer in real time, as it takes ~40s to process a single input image.

To address the latency challenge, the same authors proposed Fast R-CNN [6]. In Fast R-CNN, a CNN is used to generate convolution feature maps of the input images rather than for feature extraction. The feature maps are used for ROI identification and the ROIs are warped to squares using a pooling layer, which is further transformed into a vector to be fed into a fully connected (FC) layer. The feature vector from the FC layer is used for object class prediction in the ROI, using a softmax layer, and a bounding box regressor is used for coordinate prediction. Fast R-CNN exhibited inference speeds around ~2s, significantly faster than R-CNN, but the latency is still high, making it unusable in a real-time scenario.

Faster R-CNN [7] tackled the high latency caused by the region proposal mechanism in both the prior R-CNN works, by directly feeding the image to the CNN and letting the CNN learn to perform ROI prediction. This remarkably reduced the latency to ~0.2s.

Despite their desirable accuracies, two-stage detectors are bulky and have best case latencies in the hundreds of milliseconds on high-end GPUs. These latencies and resource overheads make them impractical for embedded real-time use cases, such as in AVs.

*Single-stage detectors:* Single-stage detectors are much faster than two-stage detectors because they use a single feed-forward network without any intermediate stage for ROI proposals. The YOLO algorithm was revolutionary when it came out in 2016 as it reframed object detection as a single-stage regression problem, from image feature extraction, to bounding box generation, and object classification. The follow-up variants of YOLO made it faster and more accurate while preserving the single shot detection philosophy. YOLOv4 introduced two important techniques: 'bag of freebies' (BoF) which involves improved methods for data augmentation and regularization during training and 'bag of specials' (BoS) which is a post processing module that leads to better mAP and faster inference [26]. YOLOv5 [8] proposed additional data augmentation and loss calculation improvements. It also used auto-learning bounding box anchors to adapt to a given dataset. Even though YOLO models provide good inference speed, they have a class imbalance problem when detecting small objects. This issue was addressed in the RetinaNet [9] detector that used a focal loss function during training and a separate network for classification and bounding box regression. Table 1 shows a comparison of various two-stage and single stage object detectors on the COCO dataset [27].

TABLE 1: Metrics comparison of two-stage vs single-stage detectors

| Name | Type | mAP | Inference rate (fps) |
|---|---|---|---|
| R-CNN [4] | two-stage | 42% | 0.02 |
| Fast R-CNN [5] | two-stage | 19.7% | 0.5 |
| Faster R-CNN [6] | two-stage | 78.9% | 7 |
| RetinaNet [8] | single-stage | 61.1% | 90 |
| YOLOv4 [24] | single-stage | 65.7% | 62 |
| YOLOv5 [7] | Single-stage | 56.4% | 140 |

While single-stage detectors are faster than two-stage detectors, they still incur significant inference times when deployed on an embedded board. To further reduce latency, model compression techniques, such as pruning, quantization, and knowledge distillation, are essential to consider. Quantization requires specialized hardware support for efficient deployment, which may not be available in embedded boards. Knowledge distillation requires the student model to be robust in order to absorb and retain the information from the teacher model, which requires both time and complex computation. Compared to its counterparts, pruning is neither computationally complex nor hardware bound, so we focus on pruning for accelerating object detector inference in this work.

*B. DNN Model Pruning*

Pruning an object detection model aims to reduce model footprint and computational complexity by removing weight parameters from the model using some criteria. Consider a deep learning model with $L_n$ number of layers. The most compute-intensive operation of a deep learning model is the Convolution (Conv) layer. If each Conv layer has $k_n$ number of kernels with $W_n$ number of non-zero weights, during inference, the computational cost of the model is a function of $(W_n \times k_n \times L_n)$. This computational cost increases dramatically as the parameters involved increases, as is the trend in modern deep learning models. By performing parameter pruning, we can induce sparsity in the model which will decrease the parameters in $W_n$ and through kernel pruning we can also decrease $k_n$. This decreases the overall computational cost. Emerging computing platforms provide software compression techniques [28] which can compress the input and weight matrices in response to the presence of zero valued (pruned) parameters, thus skipping them entirely during model execution. The skipping operation may optionally also be performed by the hardware with specifically designed hardware [29].

Pruning approaches from prior work can be classified into three major categories: unstructured pruning, structured pruning, and semi-structured or pattern-based pruning.

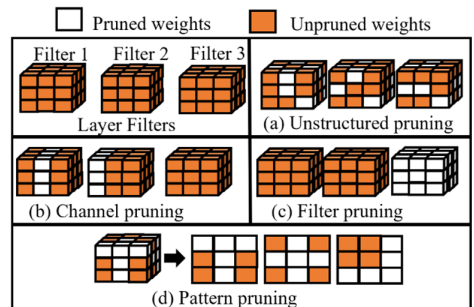

Fig. 1: Illustration of different pruning methods

*Unstructured pruning:* In unstructured pruning, redundant weights (Fig 1(a)) are pruned opportunistically, while keeping the loss to minimum which helps in retaining the accuracy of the model. Several unstructured pruning schemes have been proposed, such as: weight magnitude pruning, that focuses on replacing a set of weight below a predefined threshold to zero [13], [14]; gradient magnitude pruning, that prunes a set of weights whose gradients are below a predefined threshold [15], [16]; synaptic flow pruning, which is an iterative pruning technique that uses a global scoring scheme and prunes a set of weights until the global score drops below a threshold [17]; and second order derivative pruning, that calculates the second order derivative of weight by replacing a set of weights by zero and keeping the loss of the network close to the original loss [18]. These approaches negatively impact thread level parallelism due to the load imbalance caused by different level of sparsity from different weight matrices. Irregular sparsity also affects memory performance due to changes it creates in data access locality, leading to reduced benefits from caching across various platforms (GPUs, CPUs, TPUs).

*Structured pruning:* In structured pruning, an entire filter (Fig. 1 (c)) [19]-[21] or consecutive channels (Fig. 1(b)) [22], [23] are pruned, to increase sparsity of the model. Filter/channel pruning provides a more uniform weight matrix and reduces the size of the model. The reduced matrix help in reducing the number of multiply and accumulate (MAC) operations compared to that of unstructured pruning. However, structured pruning also decreases the accuracy of the model since weights that can be contributing to the overall accuracy of the model will also be pruned along with the redundant weights. Structured pruning can also be used with acceleration algorithms like TensorRT [24]. Unlike unstructured pruning, due to the uniform nature of the weight matrix, structured pruning can better utilize the hardware acceleration provided by various platforms in terms of memory and bandwidth [21], [23]

*Semi-structured pruning:* Semi-structured pruning, also called pattern pruning, is a combination of structured pruning and unstructured pruning schemes (Fig. 1(d)). This type of pruning utilizes kernel patterns that can be used as a mask on a kernel. A mask prevents the weights it covers from being pruned, inducing partial sparsity in a kernel. By evaluating the effectiveness of the pruned kernel, by utilizing L2 norm for example, the most effective pattern masks can be identified and deployed during inference. Since the kernel patterns can only prune a fixed number of weights inside a kernel, they will induce lesser sparsity than that of its counterparts [30], [31]. To overcome this issue, pattern pruning is applied together with connectivity pruning which prunes some of the kernels entirely. However, most modern object detectors have a large number of 1×1 kernels which contain redundant weights that are not pruned during this process. This is because, pattern pruning techniques typically focus on kernels with sizes 3×3 and above, that have more candidate weights for pruning. Connectivity pruning also reduces the accuracy of the model since several important weights in a particular kernel are also removed during this process. However, kernel pattern pruning due to its semi-structured nature can still leverage hardware parallelism to reduce inference times of the model [31].

## III. MOTIVATION

Object detectors designed for use in AVs require high accuracy, but consequently these models also have overheads such as a large memory footprint and higher inference time [38]. To overcome these issues, we need to come up with a model that can be lightweight and can achieve high accuracy. Single-stage detectors such as YOLOv5, RetinaNet, Detection Transformer (DETR), and YOLOR are a good starting point to achieve real-time object detection goals, but these models still have a high memory footprint which can decrease model performance. Table 2 summarizes the inference time as the size of the object detector model increases, on a Jetson TX2.

TABLE 2: Comparison of model sizes vs. execution time

| Models | Number of parameters (Millions) | Execution time (sec) |
|---|---|---|
| YOLOv5 [8] | 7.02 | 0.7415 |
| YOLOX [11] | 8.97 | 1.23 |
| RetinaNet [9] | 36.49 | 6.8 |
| YOLOv7 [34] | 36.90 | 6.5 |
| YOLOR [10] | 37.26 | 6.89 |
| DETR [32] | 41.52 | 7.6 |

In order to reduce latency of operation, while retaining model accuracy a pruning technique can be employed. Among pruning techniques, pattern-based semi-structured pruning can offer better sparsity over unstructured pruning, while ensuring better accuracy than structured pruning techniques. Semi-structured pruning also allows for more regular weight matrix shapes, allowing the hardware to better accelerate the model inference. Simultaneously, it does not prune entire kernel weights, unlike structured pruning, thus retaining more information and hence ensuring better accuracies. Hence, pattern-based pruning techniques can generate models with high sparsity and high accuracy, ideally.

However, a caveat of pattern-based pruning, which limits the achievable sparsity and hence the inference acceleration benefits, is that current techniques primarily focus on 3×3 kernels. Most state-of-the-art models such as YOLOv5, RetinaNet and DETR consist of 68.42%, 56.14% and 63.46% of small 1×1 kernels, respectively. So, to increase the sparsity of such models, pattern-based pruning techniques sometimes employ connectivity pruning on these 3×3 kernels [30]. But the 'last kernel per layer' criteria used in connectivity pruning contributes to loss of important information which can affect the accuracy of the model. So, we elect to avoid connectivity pruning in our pruning framework. Moreover, this technique still does not address the 1×1 kernels, which constitute a significant portion of the kernels, as mentioned above.

To address these shortcomings, we propose a three-step pruning approach to prune 1×1 kernels: 1) group 1×1 kernels to form 3×3 temporary weight matrices; 2) apply kernel pattern pruning on these weight matrices; 3) decompose the temporary weight matrices to 1×1 kernels and reassign to their original layers. Our approach thus increases the sparsity of the model while preserving important information which contributes to the accuracy of the model.

## IV. *R-TOSS* PRUNING FRAMEWORK

In this section, we describe our novel *R-TOSS* pruning framework and detail how we have implemented the previously mentioned improvements to the kernel pruning technique on the YOLOv5 and RetinaNet object detectors. A straightforward approach to pruning, while retaining much of the original performance of the model, is to adopt an iterative pruning approach. But this is a naïve approach as an iterative approach can quickly become unwieldy in terms of computational cost and time requirement as the model sizes increase. As mentioned in Section III.C, the model sizes of modern object detectors are increasing, but for many application spaces which employ them, such as AVs, their accuracy cannot be compromised. Our *R-TOSS* framework (Fig. 2) adopts an iterative pruning scheme with several optimizations for reducing computational cost and time overheads. We start by using a depth first search (DFS) algorithm which is used to find the parent-child layer couplings within the model. The parent-child graphs thus obtained are used to reduce the computation requirements for pruning. The reduction in computation costs happen as the pruning at the parent layer is reflected in its child layers within the graph. We follow up DFS with identifying the 3×3 and 1×1 kernels within the sub-graphs and applying kernel size specific pruning to them. These algorithms are discussed in detail, in the following subsections.

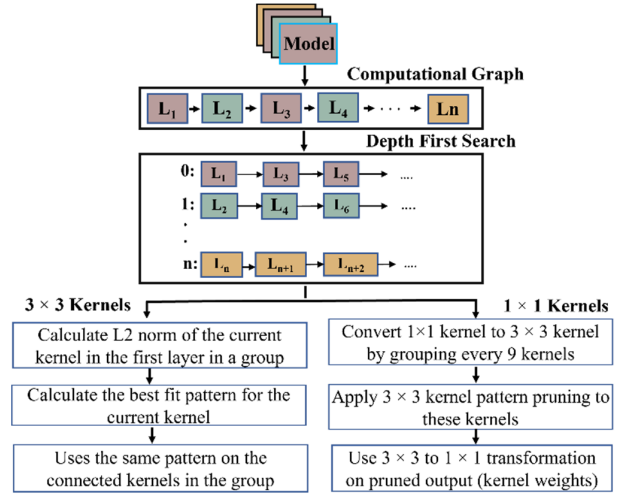

Fig. 2: An overview of the proposed *R-TOSS* pruning framework

### A. DFS algorithm

Algorithm 1 shows the pseudocode for the DFS algorithm. Using the pretrained model as input, we compute the computational graph (*G*) using the gradients obtained from backpropagation. An empty list (*group_list*) is initialized (line 2) to store the parent-child layer groups. We then traverse the model layers (*l*) and apply DFS search on the computational graph *G* to identify the parent of that layer. If a layer does not have any parent layer, then we assign that layer as its own parent layer ($l_p$) (lines 7 - 9) and this becomes a group. If a layer is identified as the child layer ($l_c$) to any layer in *group_list* (line 5) then this layer now becomes the parent layer ($l_p$) of the child layer ($l_c$) and added to that group (lines 5 - 6). Each parent layer ($l_p$) can have multiple child layers ($l_c$) but each child layer can only have one parent layer ($l_p$). This process continues until all the layers are assigned to a group. Since layers in each group have coupled channels in them, they also share their kernel properties, hence they can share the same kernel patterns.

### B. Selecting kernel patterns

We generate pattern masks in all possible combinations via standard combinatorics, using the following equation:

$$n(k) = \binom{n}{k} = \frac{n!}{k!(n-k)!} \quad (1)$$

where, *n* is the size of the matrix and *k* is the size of the pattern mask. We then narrow down the number of kernel patterns used, using the following two criteria: 1) we drop all patterns without adjacent non-zero weights; this is done to keep the semi-structured nature of the kernel patterns; 2) we select the most used kernel pattern by calculating the L2 norm of the kernel using random initiations in the range [-1, 1]. The value of k can range from 1 to 8, which can generator 8 different type of pattern groups. To increase the sparsity level of the model, the number of non-zero weight in a pattern should be lower. Prior work [20], [30] on kernel pattern pruning has used 4-entry patterns that consist of 4 non-zero weights in a kernel. But this leads to models with relatively low sparsity and to overcome this issue the authors of these works have utilized connectively pruning. Due to the drawbacks of connectivity pruning discussed in section II, we propose to use 3-entry pattern (3EP) and 2-entry pattern (2EP) kernel patterns, which uses 3 and 2 non-zero weights respectively, in our *R-TOSS* framework.

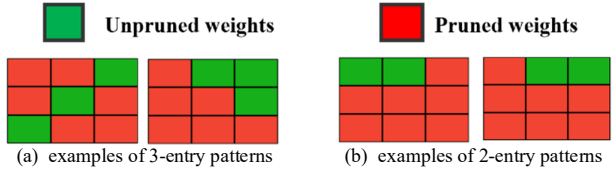

(a) examples of 3-entry patterns  (b) examples of 2-entry patterns
Fig. 3: Illustration of kernel patterns

**Algorithm 1:** *Layer grouping using DFS*

**Inputs:** *Pretrained model (M)*
1: *compute the computational graph (G) of the pretrained model*
2: *group_list ← Ø*
3: **for** *l in M:*
4: | *apply DFS with G and find parent [$l_p$] of $M[l_c]$*
5: | **if** *group_list[$l_p$[0]]* **then**
6: | | *group_list[$l_p$] ← $M[l_c]$*
7: | **else**
8: | | *group_list[$l_p$] ← 0*
9: | | *group_list[$l_p$] ← $M[l_c]$*
10: | **end**
11: **end**

**Output:** *a list of parent-child layer groups (group_list)*

### C. 3×3 kernel pruning

Algorithm 2 shows the pseudocode of the 3×3 kernel pattern pruning using the proposed kernel patterns, examples of which are illustrated in Fig 3. We start by using the 3×3 parent kernels weights ($K_W$) from Algorithm 1 as input and initializing a variable (*shape*) to store the shape of the kernel weights (line 1). We also create a pattern dictionary (*kernel_patterns_dict*) consisting of 3EP (Fig 3(a)) and 2EP (Fig 3(b)) patterns (line 3). We then traverse the 3×3 kernels and store the weight matrices of the current 3×3 kernel in the layer as *temp_kernel* (line 5). We then initialize an empty list (*L2_dict*) that can store the L2 norm of the *temp_kernel* after applying the kernel pattern from the pattern dictionary. We then iterate through the kernel pattern in the *kernel_patterns_dict* and calculate the L2norm of the kernel after applying the kernel pattern. This L2norm is stored in the *L2_dict* list along with the key of the current pattern from the *kernel_patterns_dict* (lines 7-10). We then find the best kernel pattern for the *temp_kernel* by using the L2norm value from the *L2_dict* and store the index of the kernel pattern in the *bestfit* variable (line 11). The index from *bestfit* is now used as the kernel pattern for the kernel and updated to its original weight matrices $K_W$ (lines 12-14). We then iterate through all the kernels in the parent layer and store this as the kernel mask for the rest of the 3×3 kernels in the parent layer group ($l_P$) from Algorithm 1. Once suitable patterns for the parent kernels are found, those patterns are also applied to the corresponding children, by utilizing the convolution mapping. We also apply this pattern matching approach to the 1×1 kernels by performing a 1×1 to 3×3 kernel transformation (see Section IV.D). Since we apply the same kernel mask to all the kernels in a particular group, we can reduce the time taken by the framework to prune the entire model. From experiments, we reduced the total number of patterns required to 21 patterns. Since we have only 21 pre-defined kernel patterns at inference, the kernels with similar patterns are grouped together, which can reduce the overall computational cost and speed up inference.

### D. 1×1 kernel transformation

By performing 1 × 1 to 3 × 3 transformation we remove connectivity pruning from kernel pruning. This can ensure we can maintain the accuracy of the model and mitigate losses that arise from connectivity pruning. 1×1 kernel pruning can also speedup inference by grouping similar kernel patterns together. Algorithm 3 shows the pseudocode for performing 1×1 kernel pruning. We start by using 1×1 kernel weights $K_w$ from the parent layer from Algorithm 1 (*group_list*) as input. We then initialize a list *FL* that is used to store the flattened 1×1 kernel weights from $K_w$ (lines 1-2). Subsequently, a *temp_array* for storing the temporary weight matrices is initialized. We iterate through the flattened array *FL* and group every 9 weights in the list into 3×3 temporary weight matrices that are stored in *temp_array* (lines 5-11). This process continues till we reach the end of the list or if the values are less than 9. At this point the left-over weights are considered as zero weights and pruned (line 13). We then use Algorithm 2 to perform 3×3 kernel pruning with the temporary 3×3 weight matrices from *temp_array* (line 14). The output matrices from Algorithm 2 are stored back into *temp_array* which is transformed back into 1×1 kernels and appended back into the original 1×1 kernel weights (lines 15-16).

**Algorithm 2:** *3×3 Kernel Pruning*

**Inputs:** *3×3 Kernel Weights ($K_W$)*
1: *shape ← $K_W$.shape*
2: *kernel_patterns_dict ← patterns used to prune the kernel.*
3: **for** *i in range (shape[0])* **then**
4: | **for** *j in range (shape[1])* **then**
5: | | *temp_kernel = $K_W$[i, j, :, :].copy()*
6: | | *L2_dict*
7: | | **for** *key, pattern in kernel_patterns_dict.items()* **then**
8: | | | **for** *index in pattern* **then**
9: | | | | *L2_norm = L2.norm(temp_kernel)*
10: | | | | *L2_dict[key] ← L2_norm*
11: | | | | *bestfit ← best fit pattern for the kernel in terms of L2norm*
12: | | | | **for** *index in patterns_dict1[bestfit]* **then**
13: | | | | | *$K_W$[i, j, index[0], index[1]] = 1*
14: | | | | **end**
15: | | | **end**
16: | | **end**
17: | **end**
18: **end**

**Outputs:** *Pruned 3×3 kernels.*

**Algorithm 3:** *1×1 Kernel Pooling*

**Inputs:** *1×1 Kernel Weights ($K_W$)*
1: *FL ← Ø*
2: *FL = [w for k in $K_W$ for w in K]*
3: *temp_array ← Ø*
4: *$K'_W$ ← Ø*
5: **for** *i in range (0, length(FL), 9):*
6: | *t1 ← Ø*
7: | *t1 = FL [i : i+9]*
8: | **if** *t1.shape[0] == 9* **then**
9: | | *t1 = t1.reshape(3,3)*
10: | | *L2_norm_array.append(L2norm(t1,2))*
11: | | *temp_array.append(t1)*
12: | **else**
13: | | *temp_array.append(t1=0)*
14: | *Apply 3×3 kernel pruning on temp_array using Algorithm 2*
15: | *temp_array = output weight matrices from Algorithm 2*
16: | *Reshape temp_array to 1×1 and append to $K_W$*
17: **end**

**Outputs:** *pruned 1×1 kernel*

## V. EXPERIMENTS AND RESULTS

In this section, we evaluate our proposed *R-TOSS* framework on Nvidia RTX 2080Ti and Jetson TX2 platforms in terms of sparsity, mAP, accuracy, and energy usage of the model pruned using our framework and compare it to state-of-the-art pruning techniques.

### A. Experimental setup

Our framework is implemented on YOLOv5s, which is a smaller variant of the well-known YOLOv5 with 25 layers and 7.02 Million parameters [8] and RetinaNet that consists of 186 layers and 36.49 million parameters [9]. We implemented object detectors with *R-TOSS* using Python and Pytorch and trained them on an NVIDIA RTX 2080Ti GPU [36]. The trained framework is then evaluated using the RTX 2080Ti GPU and also deployed on a Jetson TX2 [37] embedded AI computing device. We use the KITTI automotive dataset [39], with an input frame size of 640×640, and a split of 60:40, for training and inference, respectively. The KITTI dataset is a widely used dataset comprised of traffic scenes, making it ideal for AV perception model training. We measure inference times across models in terms of milliseconds (ms) and the mAP with an IoU threshold of 0.5 AP@[.5:.95].

TABLE 3: Table showing sensitivity analysis of *R-TOSS* framework in terms of induced sparsity, mAP, and inference time for YOLOv5s and RetinaNet

| Entry pattern Vatiant | YOLOv5s | | | | RetinaNet | | | |
|---|---|---|---|---|---|---|---|---|
| | *Reduction ratio* | *mAP* | *Inference Time (ms)* | *Energy Usage (J)* | *Reduction ratio* | *mAP* | *Inference Time (ms)* | *Energy Usage (J)* |
| *R-TOSS* (5EP) | 1.79× | 72.6 | 11.09 | 0.97 | 1.45× | 66.09 | 157.24 | 14.27 |
| *R-TOSS* (4EP) | 2.24× | 70.45 | 10.98 | 0.91 | 1.6× | 75.8 | 150.58 | 13.62 |
| *R-TOSS* (3EP) | 2.9× | 78.58 | 6.9 | 0.478 | 2.4× | 79.45 | 72.98 | 6.45 |
| *R-TOSS* (2EP) | 4.4× | 76.42 | 6.5 | 0.454 | 2.89× | 82.9 | 64.83 | 5.50 |

### B. Sensitivity analysis on R-TOSS pruning framework

We performed a sensitivity analysis study on our *R-TOSS* framework to determine the impact of considering different sizes of kernel patterns. Table 3 shows the results of the study, performed on the RTX 2080Ti. We explored 4-entry patterns (4EP) and 5-entry patterns (5EP), with 4 and 5 nonzero weights in a kernel, respectively, along with our 3EP and 2EP patterns discussed earlier. From the results it can be seen that while the 2EP pattern performs better in terms of sparsity induced, inference time, and energy usage on YOLOv5s, it has lesser mAP than when using the 3EP pattern. We can also observe that 2EP performs better in terms of sparsity induced, mAP, inference time, and energy usage on the RetinaNet model. The performance improvement in terms of inference time and energy usage is due to the higher achieved sparsity of the models. The results indicate that that our proposed 3EP and 2EP kernel patterns can provide faster and also more accurate results than that with the 4EP and 5EP variants of *R-TOSS*. In the next subsection, we compare the performance of 3EP and 2EP variants of *R-TOSS* with other state-of-the-art pruning frameworks.

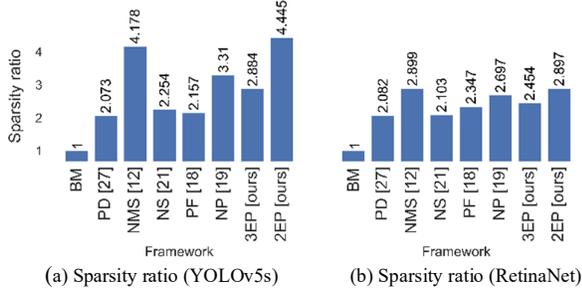

(a) Sparsity ratio (YOLOv5s)   (b) Sparsity ratio (RetinaNet)
Fig. 4: Comparison of sparsity achieved using different frameworks.

### C. Comparison results with other pruning frameworks

We compared the *R-TOSS*-3EP and *R-TOSS*-2EP variants of our proposed framework with a Base Model (BM) which does not use any pruning, PATDNN (PD) [30] which is a pattern-based pruning technique that uses a 4 entry pattern on 3×3 kernels along with connectivity pruning to increase sparsity, Neural Magic SparseML (NMS) [14] which is an unstructured weight pruning approach that uses the magnitude of the weights in a layer, with the weights below a threshold being pruned, Network Slimming (NS) [23] which uses channel pruning in which a channel is pruned based on a scaling factor for the channel in a layer, Pruning filters (PF) [20] which performs filter granularity weighted pruning, where the total sum of filters weights is calculated and filters below a corresponding threshold are pruned [20], Neural pruning (NP) [21] which uses a combination of filter pruning along with unstructured weight pruning where L1 norm is used to perform weight pruning and L2 regularization is used to perform filter pruning.

Fig. 4 shows the comparison of the sparsity ratio with other pruning frameworks from prior work, with results normalized to the baseline model BM. It can be observed that our proposed *R-TOSS*-2EP framework achieves very high sparsity across both object detector models. We were able to achieve 2.9× and 4.4× compression on the YOLOv5s model with *R-TOSS*-3EP and *R-TOSS*-2EP, respectively. Similarly, a 2.4× and 2.89× improvement in compression ratio was achieved for RetinaNet with R-TOSS 3EP and R-TOSS 2EP, respectively. Fig. 5 shows the mAP comparison. One can observe that *R-TOSS*-3EP and *R-TOSS*-2EP were able to achieve an mAP of 79.45 % and 82.9% on RetinaNet which is 8.06% and 10.98% better than the best performing framework (NMS). For YOLOv5s, the 3EP *R-TOSS* framework variant was outperformed slightly by the PD framework.

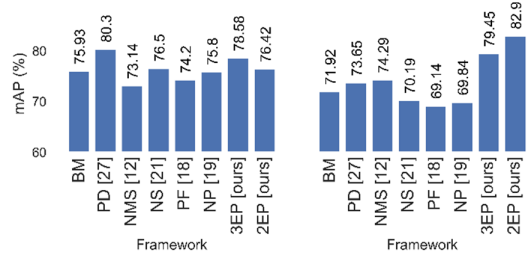

(a) mAP comparison (YOLOv5s)   (b) mAP comparison (RetinaNet)
Fig 5: Comparison of mAP achieved using different frameworks.

Our inference time results in Fig. 6 show that on RTX 2080 Ti, *R-TOSS*-3EP and *R-TOSS*-2EP were able to achieve a 1.86× and 1.97× speedup in execution time for YOLOv5s and a 1.87× and 2.1× speedup on RetinaNet compared to BM. We also outperform the best performing prior work framework (PD) by 8% and 13.3% for YOLOv5s and 43.3% and 49.6% for RetinaNet with *R-TOSS*-2EP and *R-TOSS*-3EP, respectively. Similarly, on Jetson TX2, *R-TOSS*-3EP and *R-TOSS*-2EP were able to achieve a 2.12× and 2.15× speedup in inference time on YOLOv5s model and 1.56× and 1.87× speedup on RetinaNet compared to BM. *R-TOSS*-3EP and *R-TOSS*-2EP also outperformed PD with 2.6% and 4.27% faster execution time on YOLOv5s and 5.94% and 21.62% on RetinaNet.

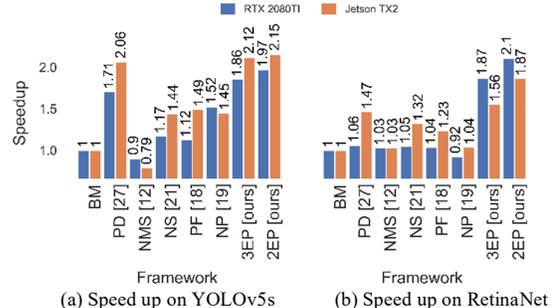

(a) Speed up on YOLOv5s   (b) Speed up on RetinaNet
Fig 6: Speedups in models after using the pruning frameworks

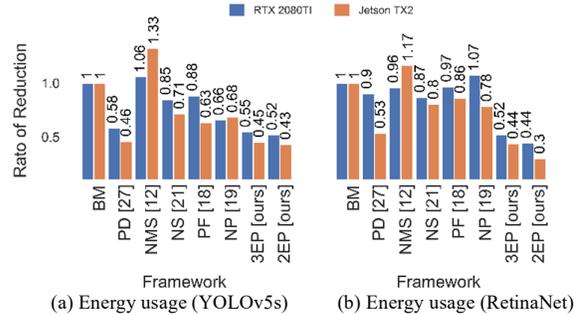

(a) Energy usage (YOLOv5s)   (b) Energy usage (RetinaNet)
Fig 7: Reduction in energy in models after using the pruning frameworks.

The models pruned using our framework also perform better in terms of energy consumption. Fig. 7 shows the comparison of reduction in use of energy among various frameworks on both

YOLOv5s and RetinaNet. For YOLOv5s, *R-TOSS*-2EP and *R-TOSS*-3EP were able to achieve 45.5% and 48.23% energy reduction over BM and a 6.5 % and 11.2% energy reduction over PD on RTX 2080Ti; and on the Jetson TX2 it achieved 54.90% and 57.01% reduction over BM and 1.84% and 6.43% reduction over PD. We also observed similar trends on RetinaNet, with 48% and 55.75% reduction of energy usage over BM and 42.46% and 50.97% reduction over PD on RTX 2080 Ti; as well as 56.31% and 70.12% reduction over BM and 18.26% and 44.10% reduction of energy usage over PD on Jetson TX2.

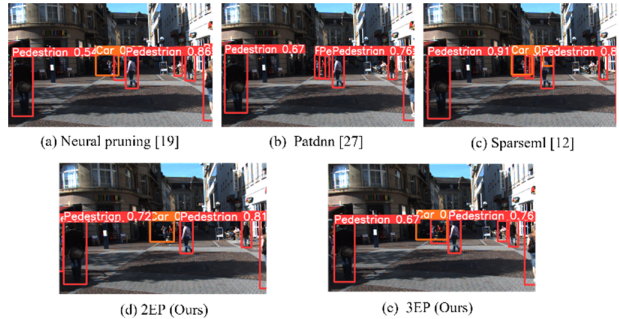

Fig 8. Comparison of inference output with other pruning techniques on KITTI automotive dataset using RetinaNet

Fig 8 illustrates the performance of different frameworks on a test case from the KITTI dataset. From the results it can be observed that *R-TOSS*-2EP especially retains the ability to detect tiny objects (the car in this example), along with better confidence scores than NP and PD. As AVs rely on fast and accurate inference to take time critical driving decision, *R-TOSS* can help achieve both speed and accuracy while keeping the energy usage lower than that of other state-of-the-art pruning techniques we have compared with.

## VI. CONCLUSIONS

In this paper we proposed a new pruning framework (*R-TOSS*) that is able to outperform several state-of-art pruning frameworks in terms of compression ratio and inference time. We were also able to increase the mAP of the object detectors compared to the mAP of the baseline models. Overall, our framework achieves significant compression ratios while improving mAP performance on two state-of-the-art object detection models, YOLOv5s and RetinaNet. The proposed framework achieves these results without any compiler optimization or additional hardware requirement. Experimental results on JetsonTX2 show that our pruning framework has a model compression rate of 4.4× on YOLOv5s and 2.89× on RetinaNet while outperforming the original model as well as several state-of-the-art pruning frameworks in terms of accuracy and inference time.

ACKNOWLEDGEMENTS

This research is supported by NSF grant CNS-2132385